# Continuous learning of spiking networks trained with local rules

D.I. Antonov[a], K.V. Sviatov[b], S. Sukhov[a],*

[a]Kotelnikov Institute of Radio Engineering and Electronics of Russian Academy of Sciences (Ulyanovsk branch), 48/2 Goncharov Str., Ulyanovsk 432011, Russia

[b]Ulyanovsk State Technical University, Severny Venets 32, Ulyanovsk 432027, Russia

E-mail addresses: d.antonov@ulireran.ru, k.svyatov@gmail.com, ssukhov@ulireran.ru

## Abstract

Artificial neural networks (ANNs) experience catastrophic forgetting (CF) during sequential learning. In contrast, the brain can learn continuously without any signs of catastrophic forgetting. Spiking neural networks (SNNs) are the next generation of ANNs with many features borrowed from biological neural networks. Thus, SNNs potentially promise better resilience to CF. In this paper, we study the susceptibility of SNNs to CF and test several biologically inspired methods for mitigating catastrophic forgetting. SNNs are trained with biologically plausible local training rules based on spike-timing-dependent plasticity (STDP). Local training prohibits the direct use of CF prevention methods based on gradients of a global loss function. We developed and tested the method to determine the importance of synapses (weights) based on stochastic Langevin dynamics without the need for the gradients. Several other methods of catastrophic forgetting prevention adapted from analog neural networks were tested as well. The experiments were performed on freely available datasets in the SpykeTorch environment.

---

* Corresponding author.
E-mail address: ssukhov@ulireran.ru (S. Sukhov)

# 1 Introduction

Humans and other animals perform a sequence of learning tasks over their lifetimes. Learning of new tasks in living organisms does not lead to the forgetting of old knowledge. This is in contrast to the modern machine learning models, which suffer from so-called catastrophic forgetting (CF): when trained on a new set of data, the artificial neural network (ANN) readily forgets all previous knowledge. Because of that, current machine-learning models typically focus on learning just a single model for a single task. Such a limitation is a major problem in many practical applications of ANNs. The methods are required for lifelong learning in ANNs that are able to accommodate new knowledge without the disruption of previously learned information (Hassabis, Kumaran, Summerfield, & Botvinick, 2017; Parisi, Kemker, Part, Kanan, & Wermter, 2019). The development of the methods intended to prevent catastrophic forgetting in ANNs is an active area of research (see (Parisi et al., 2019) for a broad review). However, the problem is still far from being solved.

The simplest solution to the problem of catastrophic forgetting is interleaved training, where multiple tasks are presented within a common training dataset (Flesch, Balaguer, Dekker, Nili, & Summerfield, 2018; Hasselmo, 2017; McClelland, McNaughton, & O'reilly, 1995). This approach may work for small datasets but is unacceptable for large ones or for continuous learning where new datasets arrive in succession. In recent years, several approaches were developed to overcome the catastrophic forgetting in analog ANNs without the necessity to keep all previous training datasets. All these approaches can be roughly divided into several groups: methods based on the peculiarities in architectures of neural networks, regularization-based methods, and memory replay methods (Hadsell, Rao, Rusu, & Pascanu, 2020). Architecture-based methods try to decrease the interference between different sub-parts of a neural network when training on new data (Schwarz et al., 2018; Yoon, Yang, Lee, & Hwang, 2017). Regularization methods try to reveal the weights important to the previous tasks and prevent their modification (Kirkpatrick et al., 2017; Zenke, Poole, & Ganguli, 2017). Memory replay methods are based on maintaining small memory of previously used training data (Chaudhry et al., 2019; Wen, Cao, & Huang, 2018) or on producing some surrogate data (Shin, Lee, Kim, & Kim, 2017; van de Ven, Siegelmann, & Tolias, 2020) that are replayed during training on new data. All these methods have some biological background. Constant neurogenesis in the hippocampus, modular organization of the brain, sparse neuronal coding diminish the interference between old and new representations and validate the architecture-based methods (Draelos et al., 2017; Yu, Migliore, Hines, & Shepherd, 2014). The synaptic persistence (synaptic plasticity) validates the regularization-based methods (Holtmaat et al., 2005; Yang, Pan, & Gan, 2009). The two-way communication between the hippocampus and neocortex validates the memory replay approach (Louie & Wilson, 2001).

The susceptibility of ANNs to catastrophic forgetting can be rooted in their differences from biological neural networks. One of the distinctions of modern ANNs from their biological counterparts is their analog nature: the interactions between nodes (neurons) are coded by analog signals. Another distinction is that the learning of ANNs is performed by a nonlocal backpropagation rule. Spiking neural networks (SNNs), the next generation of artificial neural networks, attempt to provide a more realistic model of brain functioning by taking into account the underlying neural dynamics, such as spiking, and using biologically plausible local learning rules (Ghosh-Dastidar & Adeli, 2009; Pfeiffer & Pfeil, 2018; Sanda, Skorheim, & Bazhenov, 2017; Tavanaei, Ghodrati, Kheradpisheh, Masquelier, & Maida, 2019). This potentially makes SNNs well suited to study and reveal specific mechanisms of continuous learning used by the brain. Bio-inspired learning algorithms like spike-timing-dependent plasticity (STDP), neuronal redundancy, and spike-frequency adaptation are hypothesized to be key elements for achieving continual learning (Muñoz-Martín et al., 2019; Takiyama & Okada, 2012).

This paper presents several biologically plausible approaches for continuous learning in SNNs trained with local STDP rules. The challenge of adapting the methods of analog networks for SNNs is that local training rules don't use any gradients, and the training technique does not use loss functions. Thus, besides adapting CF prevention methods from analog ANNs, we developed the original regularization method aimed at finding weights important to old knowledge.

## 2 Related work

While continuous learning is the well-researched area in analog ANNs, there are a few papers devoted to continuous learning and catastrophic forgetting in spiking neural networks.

Vaila, Chiasson, & Saxena, 2019 considered a hybrid network consisting of an STDP-trained spiking feature extractor and analog classifier (support vector machine). To prevent catastrophic forgetting in incremental class learning experiments, (Vaila et al., 2019) saved a certain number of samples from the previous dataset. In this respect, the approach of (Vaila et al., 2019) is similar to the few-shot self-reminder method (Wen et al., 2018). Another hybrid spiking-analog neural network was considered by (Vaila, Chiasson, & Saxena, 2020). The authors considered a single-incremental-task scenario of sequential learning. The authors used synaptic intelligence regularization (Zenke et al., 2017) to eliminate catastrophic forgetting. However, this regularization was applied only to the analog classifier placed on top of the spiking network. The backpropagation training of the classifier made the application of synaptic intelligence straightforward.

Other papers consider fully spiking networks without any analog add-ons. In their experiments, Ororbia, 2019 try to determine if sparse representations in SNNs help in reducing forgetting in sequential learning. For the training of their SNN, (Ororbia, 2019) used a custom learning rule based on local representations alignment. The authors found that neural processing based on spike trains indeed reduces forgetting.

The role of "sleep" in the continuous learning of spiking neural networks is discussed in a series of works from M. Bazhenov's group (Golden, Delanois, Sanda, & Bazhenov, 2020; González, Sokolov, Krishnan, Delanois, & Bazhenov, 2020). Golden et al., 2020 implemented a training phase consisting of alternating training periods on a new task and periods of "sleep." During the sleep phase, the authors changed the learning rule between the hidden and output layer from rewarded to unsupervised STDP. The sleep phase enabled spontaneous forward replay that benefited the strongest synapses. The authors claim that interleaving new task training with sleep phases enabled the network to evolve towards the configuration of the synaptic weight that supports multiple tasks. However, the approach presented in (Golden et al., 2020) requires calculating and storing average neuronal activities for the hidden layer, which is not biologically plausible. The solution could be the introduction of an additional generative network that memorizes and stores those activities.

A more elaborate model of the biological neural circuit is presented in (González et al., 2020). The network model represents a minimal thalamocortical architecture with all neurons simulated by the Hodgkin-Huxley model. The cortical connections between excitatory neurons were plastic and regulated by STDP. The authors show that sleep was able to prevent catastrophic forgetting through spontaneous reactivation (replay) of both old and new memory traces. The corresponding reorganization of synaptic connectivity allowed the consolidation of new memories and reconsolidation of the old ones.

The biologically inspired method of catastrophic forgetting prevention is discussed in (Allred & Roy, 2020). In this approach, the synaptic weights of some neurons were temporarily made more plastic by the interplay between dopamine increased learning rate and lateral inhibition. The weights of other neurons were kept more rigid that allowed to protect the previously learned information.

Wei & Koulakov, 2014 show that certain classes of STDP rules can stabilize stored memory patterns in recurrent networks despite a short lifetime of synapses. After passing through the recurrent network, the unstructured neural noise appeared to carry the imprint of memory patterns in temporal correlations. STDP, combined with these correlations, leads to reinforcement of stored patterns.

In contrast to the previous research, we consider a fully spiking network without any analog add-ons. Moreover, the SNN is trained with biologically plausible local STPD rules that exclude backpropagation and global loss function. We systematically test several methods of catastrophic forgetting prevention in incremental domain learning experiments. Besides generalizing known methods of CF prevention, we developed a new regularization-based method based on synaptic noise and Langevin dynamics. At the same time, we don't try to model the specific brain areas. Instead, in our paper, we limit ourselves to a standard feedforward architecture.

# 3 Methods

## 3.1 Spiking neural network

In our experiments, we use a feedforward convolutional spiking neural network for image classification tasks. The SNN was implemented in SpykeTorch (Mozafari, Ganjtabesh, Nowzari-Dalini, & Masquelier, 2019), an open-source simulation framework based on PyTorch. SpykeTorch supports time-to-first-spike information coding (the rank-order encoding scheme) inspired by visual processing in the brain (Thorpe, Fize, & Marlot, 1996). SpykeTorch provides non-leaky integrate-and-fire neurons with at most one spike per stimulus. The network utilizes two local learning rules: spike-timing-dependent plasticity (STDP) and reward-modulated STDP (R-STDP). R-STDP is a reinforcement learning rule in which the STDP rule is modulated by the correctness of the prediction. This behavior has a biological background and simulates the action of neuromodulators such as dopamine and acetylcholine (Brzosko, Zannone, Schultz, Clopath, & Paulsen, 2017; Frémaux & Gerstner, 2016; Legenstein, Pecevski, & Maass, 2008). The general formula that describes the modification of weights $w_{ij}$ for both rules (STDP and R-STDP) can be written as follows (Mozafari, Ganjtabesh, Nowzari-Dalini, Thorpe, & Masquelier, 2019):

$$\Delta w_{ij} = \delta_{ij}(w_{ij} - w_{\min})(w_{\max} - w_{ij}), \quad (1)$$

$$\delta_{ij} = \begin{cases} \alpha\phi_r a_r^+ + \beta\phi_p a_p^- & \text{if } t_j - t_i \leq 0 \\ \alpha\phi_r a_r^- + \beta\phi_p a_p^+ & \text{if } t_j - t_i > 0, \text{ or neuron } j \text{ never fires} \end{cases} \quad (2)$$

Here $i$ and $j$ refer to the post- and pre-synaptic neurons, respectively, $\Delta w_{ij}$ is the amount of weight change for the synapse connecting the two neurons, $(w_{ij} - w_{\min}) \times (w_{\max} - w_{ij})$ is a stabilizer term that slows down the weight change when the synaptic weight $w_{ij}$ is close to the lower $w_{\min}$ or upper $w_{\max}$ bounds. Parameters $a_r^+$, $a_r^-$, $a_p^+$, and $a_p^-$ scale the magnitude of weight change. The learning rule (2) describes the simplified STDP rule that only needs the sign of the spike time difference and uses an infinite time window. In the case of STDP, $\phi_r = 1$, $\phi_p = 0$, $\alpha = 1$, and $\beta = 0$. For R-STDP, the values of $\alpha$ and $\beta$ depend on the reinforcement signal. If a "reward" signal is generated, then $\alpha = 1$ and $\beta = 0$; in a case of "punishment," $\alpha = 0$, $\beta = 1$. Also, $\phi_r = N_{\text{miss}}/N$ and $\phi_p = N_{\text{hit}}/N$, where $N_{\text{hit}}$ and $N_{\text{miss}}$ denote the number of samples that are classified correctly and incorrectly over the last batch of $N$ input samples, respectively. The plasticity rules (1), (2) are applied for each image, while parameters $\phi_r$ and $\phi_p$ are updated after each batch of samples.

The overall organization of the spiking network was the same as in (Mozafari, Ganjtabesh, Nowzari-Dalini, & Masquelier, 2019). In particular, the network consisted of three convolutional layers (see further details in Section 4 and in (Mozafari, Ganjtabesh, Nowzari-Dalini, & Masquelier, 2019)). Each convolutional layer contained several two-dimensional grids of integrate-and-fire (IF) neurons that constitute the feature maps. Every input image was first convolved with difference of Gaussian (DoG) filters at various scales. After passing the image through DoG filters, a spike-wave was generated by an intensity-to-latency encoding (Gautrais & Thorpe, 1998). The spikes were propagated through the network till the output layer. According to the rank-order encoding scheme, the neuron in the last layer with the earliest spike time or maximum potential indicated the network's decision.

## 3.2 Continuous learning

The process of continuous learning can be described as follows. We train the naïve network on Task 1. After this initial training, we believe that the training dataset corresponding to Task 1 is no longer available (in memory replay approaches, we may keep a small set of samples of dataset 1). The pretrained network is then trained on Task 2 with techniques aimed at maintaining the memory about Task 1. After the training, the network is tested for its ability to remember Task 1 and its ability to acquire new knowledge (Task 2).

As described in the introduction, all the methods for continuous learning can be divided into three groups: architecture-based methods, memory replay methods, and regularization methods. In this section, we present biologically inspired methods of every group generalized for spiking neural networks.

### 3.2.1 Architecture-based methods

In the mammalian brain, information is represented by the sparse activity of neurons. When properly organized, this sparse activity creates a decorrelation in activity patterns and minimizes the interference between different learned tasks (Yu et al., 2014). It is hypothesized that overlapped representations play a crucial role in catastrophic forgetting; reducing this overlap would result in reduced interference (French, 1999). Sparsity in neuronal activity can lead to the emergence of independent modules inside a neural network without the explicit requirement of the predefined modular architecture of SNN (Hadsell et al., 2020). In this respect, using sparseness to combat catastrophic forgetting can be attributed to architecture-based methods.

One of the mechanisms that ensure sparse neural activity in neural networks is lateral inhibition. In neurobiology, lateral inhibition describes the process where an activated neuron reduces the activity of its weaker neighbors. The ability of lateral inhibition to prevent catastrophic forgetting in analog ANNs is demonstrated in (Aljundi, Rohrbach, & Tuytelaars, 2018). To study the effect of lateral inhibition on continuous learning, we introduce local lateral inhibition into our SNN model.

### 3.2.2 Memory replay methods

*Pseudo-rehearsal*. In his paper (Robins, 1995), Robins suggested that catastrophic forgetting can be diminished by replaying noise patterns to a network when learning a new task. In this approach, a pseudo-dataset was created by passing noise patterns through the original network and recording the network's output. This pseudo-dataset was further supplied to the neural network together with new data, and this partially helped in eliminating catastrophic forgetting (Robins, 1995).

Using white noise as pseudo-patterns works for simple networks and simple datasets. However, it fails for deep convolutional networks (Craig Atkinson, McCane, Szymanski, & Robins, 2021), where random noise has a structure quite different compared to actual input images. In our experiments, we use pseudo-patterns made of randomly oriented bars (Figure 1). Such a shape was inspired by several considerations. First, randomly oriented bars resemble the basic features (strokes) of the images of hand-written symbols (digits and letters) used in our experiment. Second, randomly oriented bars essentially represent a basis of two-dimensional Fourier transform and can, in principle, encode any image. Finally, the inspiration also came from Hubel and Wiesel experiment (Hubel & Wiesel, 1962), in which they showed that a cat's neurons in the primary visual cortex are tuned to simple features, bars.

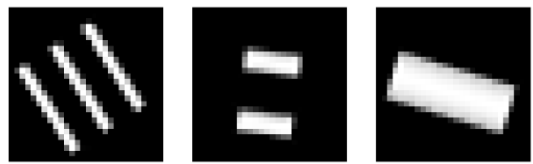

**Figure 1.** Typical patterns used for pseudo-rehearsal experiments.

As in the original Robins experiment (Robins, 1995), the pseudo-patterns consisting of randomly oriented bars (Figure 1) were first propagated through the pretrained SNN to get labels. This pseudo-dataset was later used along with new data in a continuous learning experiment. It is assumed that this surrogate dataset with a structure similar to the structure of initial data makes it possible to avoid the forgetting process by recalling previous knowledge during retraining.

*Maintaining episodic memory*. Rather than storing all training patterns as it is done during joint training on old and new datasets, one can maintain a small memory of samples (anchors, exemplars) that represent the key features of previous tasks. This simple approach of maintaining small episodic memory demonstrated superiority against many modern methods for the catastrophic forgetting prevention in analog ANNs (Chaudhry et al., 2019; de Masson d'Autume, Ruder, Kong, & Yogatama, 2019; Wen et al., 2018). The SNN network trained by R-STDP is claimed to perform well even if one reduces the number of training samples (Mozafari, Ganjtabesh, Nowzari-Dalini, Thorpe, et al., 2019). Thus, there is a good chance that learning with small episodic memory would work well in SNNs for the elimination of catastrophic forgetting.

In our experiments on continuous learning, we maintain a small memory of samples from previously learned dataset with samples picked at random. During continuous learning, these samples are mixed with new incoming training data to remind the neural network about previously learned tasks.

### 3.2.3 Regularization based methods

*Noise regularization.* The idea behind this approach is that subsequent learning on a new dataset introduces some sort of "noise" to the original distribution of weights. If we teach the network to withstand the noise during original training, there is a hope that this network does not react to the perturbations introduced by a new set of data.

In biology, noise is an intrinsic property of neural circuits attributed to the unreliability of synapses (Faisal, Selen, & Wolpert, 2008; Maass, 2014; McDonnell & Ward, 2011). In the theory of SNNs, noise is found to stabilize old memories in recurrent SNNs trained by the STDP rule (Wei & Koulakov, 2014). However, it is known that training with noise is equivalent to a form of regularization in which an extra term is added to the error function (Bishop, 1995). For example, the injection of Gaussian noise into synaptic weights is equivalent to L2 regularization. It was shown before for analog ANNs that L2 regularization does not help with the prevention of catastrophic forgetting (Kirkpatrick et al., 2017). In our experiments, we test if this is also the case for the spiking neural networks.

In our work, we introduce the noise through a change of learning parameters in Eq.(2):

$$\delta_{ij} = \begin{cases} \alpha\phi_r(a_r^+ + \sigma^+ N(0,1)) + \beta\phi_p(a_p^- + \sigma^- N(0,1)), & \text{if } t_j - t_i \leq 0 \\ \alpha\phi_r(a_r^- + \sigma^- N(0,1)) + \beta\phi_p(a_p^+ + \sigma^+ N(0,1)), & \text{if } t_j - t_i > 0, \text{ or neuron } j \text{ never fires} \end{cases} \quad (3)$$

where $\sigma^{\pm}$ is the noise amplitude, $N(0,1)$ stands for the standard normal distribution.

*Freezing large weights*. Current regularization methods estimate the importance of the weights necessary for maintaining old memories by calculating the derivatives of the loss function (Kirkpatrick et al., 2017; Zenke et al., 2017). The methods penalize the changes to the important weights when learning a new task (Figure 2a). However, SNNs used in our paper have neither loss function nor gradients in training. The importance of the weights (or should it be called slow or fast weight) should be determined from some local rules. We tested several ideas for finding important weights in SNNs trained by local rules.

The connections inside biological neural networks are formed both by transient and persistent synapses (Holtmaat et al., 2005). It was found that thick spines persist for months; the role of corresponding synapses is supposedly in the stabilization of memory (Holtmaat et al., 2005). In the theory of ANNs, the reactivation of the strongest weights to prevent catastrophic forgetting was studied in (Golden et al., 2020). Thus, one assumption is that the important weights are the strongest ones. One of the options for catastrophic forgetting prevention is to reduce the change of the strongest synapses (the largest weights). The simplest solution would be to "freeze" the important (the largest) weights, completely restricting any their modification. However, this approach has a danger of impairing the ability of a neural network to learn new information. In our experiments, we demonstrate that large weights represent only a small subset of the whole weights in our SNNs. Thus, freezing these weights should not completely restrict neural network learning.

*Langevin dynamics*. As was mentioned above, SNNs used in our paper have neither loss function nor easily calculated gradients. On the one hand, that prevents us from calculating the importance of the weights based on the derivative of the loss function (Kirkpatrick et al., 2017) (Figure 2a). On the other hand, the performance of trained SNN is determined by nondifferentiable accuracy; we can change the weights in certain limits without the decrease in this accuracy (Figure 2b). To find the multidimensional region of permitted weights variation, one can use Langevin dynamics. With Langevin dynamics introduced, the weights perform Brownian motion within the region of high accuracy (Figure 2b). When the weights enter the region where the network produces the wrong prediction, the R-STDP mechanism kicks in, trying to bring weights back to their optimal values. The R-STDP mechanism creates an effective potential or "force field" $U^*$ (Figure 2b) where virtual Brownian particle moves. With enough time, the Brownian motion could cover the whole area of allowed weights $D$. Sampling the trajectory of the weights, one can make a conclusion about the shape and the size of domain $D$. This process is similar to the estimation of the probability distribution of weights in Bayesian neural networks (Sukhov, Leontev, Miheev, & Sviatov, 2020; Welling & Teh, 2011). The difference is that sampling, in our case, does not try to reproduce the loss function (or some potential energy). The forces represented by the STDP mechanism can be nonconservative, and the notion of "sampling the potential energy" loses its sense.

After enough samples have been collected during Brownian diffusion of weights, one can estimate the lower and upper bounds $w_{ij}^l, w_{ij}^u$ for the allowed weights change. Thus, in our approach, we approximate the domain $D$ by multidimensional cuboid.

In practice, after the training of SNN on the first dataset till the convergence (Task 1), the Brownian dynamics was implemented by adding Gaussian noise to *all* the weights $w_{ij}$ at each step $n$ of training:

$$w_{ij}^{n+1} = w_{ij}^n + F(w_{ij}) + \sigma N(0,1). \tag{4}$$

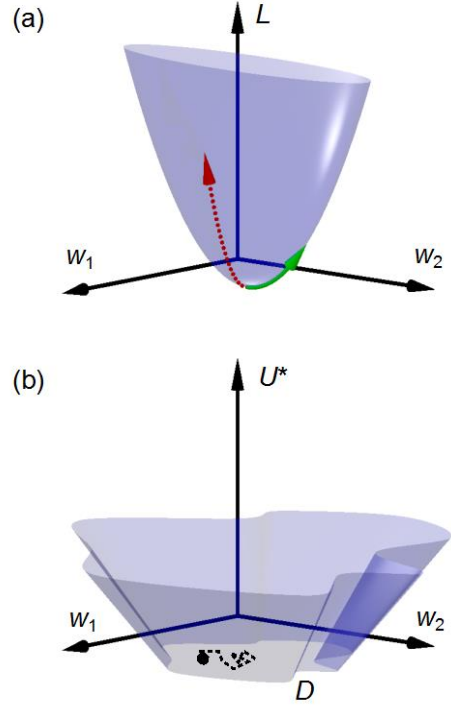

**Figure 2.** The schematics of important weights determination in analog and spiking neural networks. a) Loss function $L(w_1, w_2)$ of analog ANNs can be approximated by asymmetric paraboloid near the local minimum. The important weights are those that significantly increase the loss $L(w_1, w_2)$ after their small modification $w_i + \Delta w_i$ (red dotted curve). Change of the weights along the solid green curve results only in a minor change of the loss function.; b) SNNs trained by local rules have the domain of optimal weights $D$ that can be explored by the stochastic Brownian process (dashed line). R-STDP mechanism prevents Brownian motion from leaving domain $D$ creating "force field" with effective potential $U^*$.

Eq.(4) represents the overdamped Langevin equation for weight $w_{ij}$ with $F(w)$ being an effective "force" created by the STDP mechanism. The values of the weights were collected during $N_B$ steps of Brownian motion. The consecutive samples $w_{ij}^n$, $w_{ij}^{n+1}$ can be highly correlated (Brownian motion is known for slow mixing). Thus, one needs to collect samples only at a certain rate (every $N_{\text{demix}}$ samples). This rarified sample collection also helps in reducing computer memory size. When choosing the amplitude of noise $\sigma$, there is a compromise between the demixing rate and average accuracy maintained by the STDP process. With low noise amplitude, the STDP process is able to keep the weights well within the domain $D$. However, it might require an incredibly long time to sample the entire domain $D$ with this diffusion rate. On the contrary, large noise amplitude enables quick weights' diffusion; however, STDP cannot prevent weights from falling out of the domain $D$. In practice, we terminate sample collection during Brownian dynamics if accuracy drops below predefined minimally acceptable accuracy $A_{\min}$.

After the completion of the sampling process, the probability distribution $p(w_{ij})$ is calculated for every collected weight's value $\{w_{ij}\}$. To characterize the lower and upper bounds of weight's change (the width of $p(w_{ij})$), we estimate interdecile range $[w_{ij}^l, w_{ij}^u]$ for every $p(w_{ij})$. For lower and upper bounds $w_{ij}^l, w_{ij}^u$, we took the first and the ninth deciles of $p(w_{ij})$. The reason for using interdecile range instead of standard deviation is that probability distributions $p(w_{ij})$ are likely asymmetric and highly non-Gaussian.

The whole procedure of $w_{ij}^l, w_{ij}^u$ determination is outlined in Algorithm 1. Note that Algorithm 1 should be executed before the discarding of the initial dataset $S_1$ during continuous learning. During training on the next dataset (Task 2), the weights should be allowed to change only within intervals $[w_{ij}^l, w_{ij}^u]$.

**Algorithm 1** Finding the permitted weights range with Brownian dynamics

---
**Input:** $S_1, \sigma, N_B, N_{\text{demix}}, A_{\text{min}}$
**Output:** collected weights samples $W = \{w\}$, lower and upper bounds $[w_{ij}^l, w_{ij}^u]$ of allowed change for every weight $w_{ij}$

train neural network on Task 1 till completion
**for** $N_B$ steps
    **for** every sample in $S_1$
        make a learning step by STDP rule as in Eq. (1)
        add Gaussian noise to all the weights $w_{ij} \rightarrow w_{ij} + \sigma N(0,1)$
        store the values of the weights in $W$ every $N_{\text{demix}}$ steps
    **end for**
    terminate **for** loop if accuracy drops below $A_{\text{min}}$
**end for**
calculate probability distribution $p(w_{ij})$ for every weight
calculate lower and upper bounds $[w_{ij}^l, w_{ij}^u]$ as first and ninth deciles of $p(w_i)$

---

The range of lower and upper bounds $[w_{ij}^l, w_{ij}^u]$ has a direct relation to the importance of particular weight $w_{ij}$. The narrower is $[w_{ij}^l, w_{ij}^u]$, the more important is the weight $w_{ij}$. Thus, the presented technique may be considered as the generalization of weights-regularization methods of analog ANNs (Kirkpatrick et al., 2017).

There is some similarity of the proposed method to the method of (Golden et al., 2020). In (Golden et al., 2020), the hidden layer neurons were artificially stimulated by Poisson distributed spike trains in order to maintain spiking rates similar to that during task training. In our approach, we also supply noise to the network. However, unlike (Golden et al., 2020), we still use reinforcement learning.

# 4 Experiment

## 4.1 Initial learning of SNN

For our experiments, we used a deep convolutional SNN with a structure adapted from (Mozafari, n.d.; Mozafari, Ganjtabesh, Nowzari-Dalini, Thorpe, et al., 2019). The network had three convolutional layers (S1, S2, S3), each followed by a pooling layer (C1, C2, C3). The first and second convolutional layers consisted of 30 and 250 neuronal maps with 5 × 5 convolution-window (Mozafari, Ganjtabesh, Nowzari-Dalini, Thorpe, et al., 2019). The final third layer contained 200 neurons. The lateral inhibition radius was set to zero for all the layers. The rest of the parameters were the same as in the initial code (Mozafari, n.d.).

The input images were convolved with the difference of Gaussian (DoG) filters, locally normalized, and transformed into a spike wave. For each input image, On- and Off-center DoG filters of three different scales and with zero padding were applied (Kheradpisheh, Ganjtabesh, Thorpe, & Masquelier, 2018). Window sizes of DoG filters were set to 3 ×3, 7 ×7, and 13 ×13 (Mozafari, Ganjtabesh, Nowzari-Dalini, Thorpe, et al., 2019); the standard deviations for DoG filters were (3/9, 6/9), (7/9, 14/9), and (13/9, 26/9), respectively.

For consecutive training of the SNN, we used several datasets. One dataset, MNIST, contained images of hand-written digits (Lecun, Bottou, Bengio, & Haffner, 1998). Another dataset was a subset of EMNIST (EMNIST dataset contains hand-written letters and digits) (Cohen, Afshar, Tapson, & van Schaik, 2017). From EMNIST, we borrowed a set of 24000 hand-written capital letters arranged into ten classes (letters A, B, D, E, G, H, N, Q, R, S). To balance the data in our experiments, we also used only 24000 images from the MNIST dataset. The neurons in the output layer were divided into 10 groups (20 neurons per group) corresponding to ten classes in training data.

As in the original paper (Kheradpisheh et al., 2018), we initialized the synaptic weights with random values drawn from a normal distribution with a mean of 0.8 and a standard deviation of 0.05. At the initial moment, the learning rates were set to $a^+ = 0.004$ and $a^- = -0.003$ (Kheradpisheh et al., 2018).

At the first step, we trained our network on MNIST images (Task 1). Similar to (Mozafari, Ganjtabesh, Nowzari-Dalini, Thorpe, et al., 2019), we trained the network in a layer-by-layer manner. Layers S1 and S2 were trained by STDP for 2 and 4 epochs, respectively. The learning rates were multiplied by 2 after every 500 iterations (Mozafari, Ganjtabesh, Nowzari-Dalini, Thorpe, et al., 2019). We kept doing multiplication while parameter $a^+$ was less than 0.15, and parameter $a^-$ was larger than – 0.1125. Lower and upper bounds of weight were $w_{min} = 0.2$ and $w_{max} = 0.8$. The last layer S3 was trained for 600 epochs with R-STDP. With the proposed training technique, the network extracts regularities of the input images in lower layers S1, S2 and uses R-STDP in output layer to learn rewarding behavior.

The recognition performance of the network was tested on 4000 test samples. The experiment was performed five times, starting from different seed weights. The accuracy of classification of the test data after the training was 90.8±0.9%. For comparison, after the training similar SNN over 686 epochs (2 for the first, 4 for the second, and 680 for the last trainable layer) using the whole set of 60000 MNIST images, (Mozafari, Ganjtabesh, Nowzari-Dalini, Thorpe, et al., 2019) obtained 97.2% accuracy. The lower value of accuracy in our case is explained by the absence of lateral inhibition in layers S1 and S2. Lateral inhibition helps in better training that can be related to better feature extraction. Further effects of lateral inhibition will be discussed later.

In all our experiments, two computers were used to run the code. One PC had Intel Core i9 CPU (3.1 GHz), GeForce RTX 2080 Ti TURBO GPU with 11 Gb memory size, 16 Gb RAM, PyTorch 1.8.0, and Ubuntu 18.04. The second PC had a processor AMD Ryzen 5 (3.6 GHz), 16 Gb RAM, NVIDIA GeForce RTX 3060 GPU with 12 Gb memory, PyTorch 1.8.1, and Windows 10 Pro OS. With these computer parameters, one training epoch took approximately 2.5 minutes for 24000 training patterns.

In the second step, we intend to train our network on images of letters extracted from the EMNIST dataset (Task 2). The ultimate goal of such sequential learning is to teach the network to properly classify new patterns without losing the ability to recognize the old ones. This type of continuous learning is called incremental domain learning (Shin et al., 2017). In such experiments, the data in every dataset are arranged into the same classes but taken from different domains.

### 4.2 Catastrophic forgetting

First, we tested the ability of vanilla SNN to withstand catastrophic forgetting. Assuming that the previous MNIST dataset is no longer available, we retrained the network on EMNIST data (images of letters). The training was performed with the same set of parameters as in the case of the MNIST dataset (in particular, the learning rates $a^+$ and $a^-$ were restored to their original values). The layers S1 and S2 were again trained by STDP for 2 and 4 epochs, respectively. The last S3 layer was trained by R-STDP over 100 epochs. The training led to the relatively low accuracy in letters recognition (78.4±1.2%) and low accuracy of retaining previous knowledge (48.1±4.8%) (see Table 1 and Supplementary Figure S1(a)). Thus, coding information in spikes and using local rules did not help SNN in retaining previous knowledge during continuous learning.

**Table 1**

Average test accuracy of classification of SNN in a sequential learning experiment. Each experiment was performed 5 times. Reported is the mean (± standard deviation).

| Method | Initial training, % | Subsequent training, % | |
|---|---|---|---|
| | MNIST | MNIST | EMNIST |
| Joint training | 93.6 (±0.3) | 92.0 (±0.1) | 79.7 (±0.5) |
| Lateral inhibition | 93.6 (±0.3) | 73.3 (±0.9) | 87.3 (±0.3) |
| Pseudo-rehearsal | 93.6 (±0.3) | 43.5 (±5.0) | 79.0 (±0.7) |
| Few-shot self-reminder (0.25%) | 93.6 (±0.3) | 74.5 (±0.5) | 74.0 (±1.1) |
| Few-shot self-reminder (10%) | 93.6 (±0.3) | 91.1 (±0.4) | 77.8 (±0.6) |
| Noise regularization | 93.9 (±0.1) | 67.2 (±1.2) | 87.8 (±0.5) |
| Frozen large weights | 93.6 (±0.3) | 78.2 (±2.1) | 69.2 (±1.5) |
| Langevin dynamics | 93.6 (±0.3) | 82.2 (±0.5) | 78.3 (±1.1) |
| Catastrophic forgetting | 90.8 (±0.9)[*] | 48.1 (±4.8) | 78.4 (±1.2) |

[*] Trained over 600 epochs.

## 4.3 Catastrophic forgetting prevention

### 4.3.1 Lateral inhibition

To test the effect of lateral inhibition, we set the inhibition radius to be nonzero: inhibition radius was equal to 3 in the first layer S1, an inhibition radius was equal to 1 in the second layer S2, and inhibition radius was zero in the third layer. Lateral inhibition in the output layer S3 had no effect as one winning neuron was selected at every learning step. With these settings, the parameters of the neural network for the MNIST experiment were the same as in (Kheradpisheh et al., 2018; Mozafari, Ganjtabesh, Nowzari-Dalini, Thorpe, et al., 2019). With this new set of parameters, we redid the training of naïve SNN on Task 1 to have the accuracy approximately equal to the case with zero inhibition radius. Layers S1 and S2 were trained by STDP for 2 and 4 epochs, respectively. The last layer S3 was trained for 50 epochs with R-STDP. Even with 50 training epochs, we obtained 93.6±0.3% accuracy when testing on MNIST data.

At the next step, we continued training SNN on Task 2 (images of letters). With training parameters restored, we trained layer S1 over 2 epochs; layer S2 was trained over 4 epochs, layer S3 was trained over 100 epochs. After the training on Task 2, the accuracy of SNN on MNIST data was 73.3±0.9% with accuracy for EMNIST reaching 87.3±0.3%. In other words, one can observe gradual forgetting of previous knowledge during sequential learning. Thus, the presence of lateral inhibition resulted in graceful forgetting of old knowledge in SNN during sequential learning (see Table 1 and Supplementary Figure S1(b)).

The effect of lateral inhibition is in nonoverlapping feature representations in the first two layers. The major interference resulting in catastrophic forgetting happens in the final third layer of the network. To prove that, for a network initially trained on MNIST, we retrained its first two layers on the EMNIST dataset. The test accuracy for MNIST data dropped only by 0.3%. Thus, the first two layers with lateral inhibition turned on did not contribute to the catastrophic forgetting during continuous learning. In the rest of the experiments, we used SNN with lateral inhibition in the first two layers S1 and S2; different methods of catastrophic forgetting prevention were applied only to the output S3 layer. Correspondingly, for Task 1, we trained the naïve network only over 50 epochs to have initial accuracy similar to the case of catastrophic forgetting.

### 4.3.2 Memory replay

*Pseudo-rehearsal.* According to the pseudo-rehearsal concept, we created pseudo-patterns formed by randomly placed bars (see Figure 1). The dimensions of each pseudo-pattern corresponded to the size of the MNIST image (28 by 28 pixels). The images consisted of one to three bars with random width. In each image, bars were rotated at an arbitrary angle. The generated pseudo-patterns were fed into the SNN pretrained on the MNIST dataset and assigned with a class label. The generation of random pseudo-patterns continued until 2400 samples were collected for each class of training dataset. For a continual learning experiment, the collection of pseudo-patterns was combined with images of letters EMNIST and shuffled. The SNN was trained on this combined dataset (2 epochs for layer S1, 4 epochs for layer S2, 100 epochs for layer S3). The results of the training are shown in Table 1 and Supplementary Figure S1(c). Surprisingly, instead of fighting catastrophic forgetting, pseudo-rehearsal provoked fast unlearning of previous knowledge. This tells us that pseudo-rehearsal with random samples is unsuitable for complex convolutional SNNs.

*Few-shot self-reminder.* Fighting catastrophic forgetting by maintaining small memory of samples taken from previous datasets turned out to be very effective in analog ANNs (Wen et al., 2018). Following (Wen et al., 2018), we call this method the few-shot self-reminder. To test this concept in SNNs, we stored several samples from Task 1 (MNIST dataset) and used them as reminders during consecutive training on Task 2 (EMNIST dataset).

Two series of few-shot self-reminder experiments were performed. As the first option, we stored 6 MNIST images per class (0.25% out of 24000 total images). As the second option, 240 images per class were selected (10% of the original training dataset with 2400 images per class). The stored images were replicated to get 24000 images in the training dataset. This depleted dataset was combined with the dataset of Task 2 (images of letters) and shuffled. The pretrained model was trained on this combined dataset (2 epochs for layer S1, 4 epochs for layer S2, 100 epochs for layer S3). As a result, the recognition accuracy of previous knowledge achieved 74.5±0.5% in the case of 6 stored images per class; the accuracy reached 91.1±0.4% in the case of 2400 stored images (see Table 1). The evolution of the accuracy during training is shown in Supplementary Figure S1(d,e). Experiments show that SNN required much more stored samples per class than analog ANNs to maintain old memories (Wen et al., 2018). This conclusion somewhat contradicts the previous finding that SNN can be effectively trained on depleted datasets (Mozafari, Ganjtabesh, Nowzari-Dalini, Thorpe, et al., 2019).

### 4.3.3 Regularization-based approaches

*Noise regularization.* According to the approach discussed in Section 3.2.3, we introduced Gaussian noise to the training process. To demonstrate the effect of noise regularization, the naïve network was trained on MNIST images over 50 epochs with noise introduced as in Eq.(3). During the training, we kept the amplitude of the noise at the level of half of the average weight change $\sigma^+ = 0.5|\langle a^+ \rangle|$ and $\sigma^- = 0.5|\langle a^- \rangle|$. The noise was applied only to the weights that were being changed by STDP at a given step. Surprisingly, the high level of noise did not decrease the accuracy of the network during the initial training on MNIST data (see Table 1). Subsequent training on images of letters was done during 2 epochs for layer S1, 4 epochs for layer S2, 100 epochs for layer S3 with the same parameters of noise. The results of the training are shown in Table 1. The evolution of the accuracy during training is shown in Supplementary Figure S1(f). One can see that the noise allowed effective learning of new knowledge (images of letters). However, the noise did not help in maintaining old memories (images of digits): test accuracy on MNIST data was 67.2±1.2%.

*Freezing large weights.* In this experiment, we try to determine important weights in the network and prevent them from changing. As was discussed in Section 3.2.3, one assumption is that the largest weights (strongest synapses) are the most important. For a network to learn new information, the connectivity of the network should be sparse: trained SNN should have only a small ratio of strong synapses. In our SNN trained by local STDP and R-STDP rules, we do observe such sparse connectivity. Figure 3 shows the distribution of weight magnitudes in layer S3 of SNN trained on the MNIST dataset.

One can see that most of the weights have the value 0.2 (minimum allowed weight value preset in the code). The amount of weights with a maximum allowed value of 0.8 is the order of magnitude smaller.

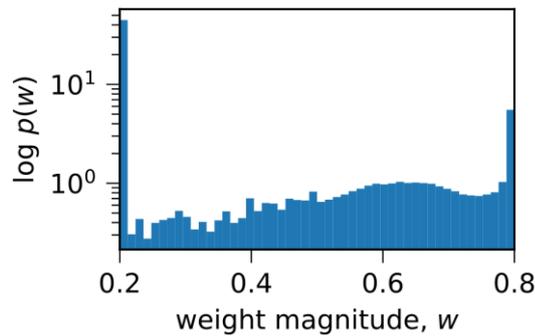

**Figure 3.** Probability distribution $p(w)$ of weight values $w$ in output layer S3 of SNN trained on MNIST dataset.

The formation of sparse connectivity in spiking neural networks was studied in (Sacramento, Wichert, & van Rossum, 2015). The authors found that sparse wiring appears as a result of an imbalance towards depression in learning rates.

To investigate the role of the largest weights on the network's performance, we pruned (set their values to 0.2) various ratios of the smallest weights in layer S3. The result of this pruning is shown in Figure 4. The initial accuracy of the network trained on the MNIST dataset (layer S1 was trained on 2 epochs, layer S2 was trained on 4 epochs, layer S3 was trained over 600 epochs) was 96.4%. Pruning 90% of the weights resulted in accuracy of 92.8%. With 95% pruned weights, the accuracy dropped to 85.3%. The accuracy of prediction significantly dropped when large weights began to be pruned (Figure 4). Thus, in terms of classification accuracy, the largest weights indeed have a major role.

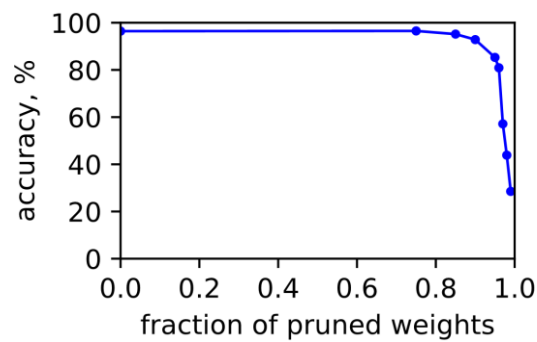

**Figure 4.** Test accuracy as a function of the fraction of pruned smaller weights.

Next, we checked if freezing the largest weights actually leads to the conservation of old knowledge. For the network pretrained on MNIST images, we froze 10% of the largest weights in the network's layer S3. Subsequent training on Task 2 (images of letters) was performed over 2 epochs for layer S1, 4 epochs for layer S2, and 100 epochs for layer S3. The test accuracy on MNIST data indeed improved to some extent (see Table 1). However, that happened at the expense of slow learning of new knowledge. Even with a small amount of completely frozen weights, the network's ability to learn was greatly impaired (see Table 1 and Supplementary Figure S1(g)).

*Langevin dynamics.* As one could see in the previous experiment, the largest weights are not necessarily the most important in terms of retaining previous knowledge. Thus, a more suitable procedure for important weights finding should be implemented. Moreover, freezing the weights prevents the neural network from acquiring new knowledge. Thus, weights should be allowed to have some variability. With Langevin dynamics described in Section 3.2.3, we attempt to determine the safe intervals for the weights change that allow SNN to learn new knowledge but also do not significantly affect the accuracy of SNN for old knowledge. After training the network on Task 1 (images on digits), we run the Brownian dynamics on weights according to Algorithm 1. Namely, we continued training SNN with additional Gaussian noise added to *all* the weights. When choosing the noise amplitude $\sigma$, there is a compromise between the time required for the weights to explore the whole domain of acceptable values and the ability of STDP to maintain the desired accuracy. We chose the amplitude of the noise $\sigma$ to be $4 \cdot 10^{-4}$. With this amplitude, it would take 100 epochs for a weight to traverse the whole range of allowed values [0.2, 0.8]. However, in the experiment with Brownian dynamics, the accuracy dropped below 80% after 25 epochs. Thus, we used only the first 25 epochs of Brownian dynamics to evaluate weights' variation intervals. Following Algorithm 1, we determined lower and upper bounds for the weights' change $[w_{ij}^l, w_{ij}^u]$ as the first and ninth deciles of corresponding probability distributions $p(w_{ij})$.

At the next step, we performed learning of SNN on Task 2 (images of letters), allowing each weight $w_{ij}$ to change only within the corresponding interval $[w_{ij}^l, w_{ij}^u]$. After the training on images of letters (2 epochs for layer S1, 4 epochs for layer S2, and 100 epochs for layer S3), SNN obtained the recognition accuracy on the previous dataset 82.2±0.5%, which was much better than the accuracy in the case of frozen weights (Table 1 and Supplementary Figure S1(h)). At the same time, the network was able to learn new knowledge with a precision of 78.3±1.1%.

## 4.4 Joint training

To have the best-case scenario of the achievable accuracy during sequential training, we also performed the training under the assumption that the network has access to the previous dataset while training on the new one (the case of joint training or offline learning). Accordingly, the datasets containing images of digits and letters were combined and shuffled. Then, the network pretrained on the MNIST dataset was continued to be trained on this combined dataset (2 epochs for layer S1, 4 epochs for layer S2, and 100 epochs for layer S3). The result of this training is shown in Table 1 and in Supplementary Figure S1(i). Unlike analog ANNs, we observe somewhat different behavior. Namely, while the classification accuracy for digits achieved 92.0±0.1%, the classification test accuracy for letters struggled to reach high values achieving only 79.7±0.5%. The performance could possibly be improved by fine-tuning training parameters in STPD and R-STPD or by employing the regularization techniques (for example, dropout). However, further improving STPD rules and fine-tuning of hyperparameters we left for future research.

# 5 Discussion

Most of the proposed techniques (except pseudo-rehearsal and regularization with noise) to some extent helped in decreasing the forgetting of old knowledge during sequential learning. The most successful approach appeared to be the few-shot self-reminder. However, in the case of spiking networks, it required much more stored samples of old data to achieve high accuracy. Future improvements for the few-shot self-reminder method can be achieved by proper choice of samples for the memory buffer (Aljundi, Belilovsky, et al., 2019; Aljundi, Lin, Goujaud, & Bengio, 2019; Chaudhry, Gordo, Dokania, Torr, & Lopez-Paz, 2021; Chaudhry et al., 2019; Isele & Cosgun, 2018; Rebuffi, Kolesnikov, Sperl, & Lampert, 2017).

The success of the few-shot self-reminder method allows to suggest that the most promising method for catastrophic forgetting prevention in SNNs would be the formation of a synthetic dataset reminiscent of the previous data with a generative approach (Shin et al., 2017; Su, Guo, Tan, & Chen, 2019). However, the generative SNNs are yet to be developed. Also, the efficiency of maintaining the old knowledge would greatly depend on the quality of synthetic data. This conclusion follows from the failure of the pseudo-rehearsal method, where even somewhat similar to the original data pseudo-samples could not prevent catastrophic forgetting. This is in line with prior findings that pseudo-rehearsal does not scale well to complex networks with large input spaces (C Atkinson, McCane, Szymanski, & Robins, 2018).

Experiments show that sparse representations created by lateral inhibition in SNN are insufficient for the complete elimination of forgetting (although lateral inhibition definitely has a positive effect). In (Aljundi et al., 2018), an additional condition for lateral inhibition in analog ANNs was the reduced inhibition from/to neurons which have high neuron importance. Such a mechanism is probably needed for SNNs as well.

Our research shows that regularization with noise helps in acquiring new knowledge (helps in generalization) but cannot prevent catastrophic forgetting. This behavior is similar to the one in analog ANNs. To have a positive effect, one could need the asymmetric STDP rules during network learning (Wei & Koulakov, 2014). Additionally, recurrent connections in the network's architecture could be needed to rehearse the stored patterns (Wei & Koulakov, 2014). In this respect feedforward architecture used in our paper is insufficient to ensure maintaining old memories.

Golden et al., 2020 show that noise replay helps in strengthening already strong synapses corresponding to old tasks. However, we demonstrated that important synapses and strong synapses in SNNs are not always the same thing.

Our experiments show that freezing even a small amount of weights (10%) leads to a severe drop in the ability to learn a new task. In this respect, providing a final range for weights variability (as in the Langevin dynamics method) gives more flexibility in training a network on new datasets. The method based on the determination of weights' acceptable domain of variation is a promising technique. In our experiments, the approach based on Langevin dynamics demonstrated the second-best result ceding only to the few-shot self-reminder method with 10% retained samples. The Langevin dynamics method can be further improved. In the current paper, we approximated the domain of acceptable weights by a cuboid. The shape might be reconstructed more precisely by taking into account the correlations in weights variation. Using the whole probability distribution $p(w)$ and not only the interdecile range may also be useful.

Because of the early termination of the diffusion process in our experiment, likely, not the whole domain of appropriate weights was sampled. We needed early termination in Langevin dynamics experiments because R-STDP could not recover the network's accuracy even with the noise turned off. Thus, the cases where SNN failed to learn on new data can sometimes be attributed to the shortcomings of the R-STDP learning algorithm. Further developments of local learning rules are still needed.

# 6 Conclusion

In this work, we performed incremental domain learning experiments for spiking neural networks trained by local training rules. We found that SNNs are prone to catastrophic interference, and coding information in spikes does not solve the problem by itself. We tested several methods for catastrophic forgetting prevention that do not require the modification of the STDP rule. Besides the methods adapted from the case of analog ANNs, we developed an original approach allowing us to determine the importance of the weights in the form of their allowed variation range. This approach can be considered as a generalization of the elastic weight consolidation method for SNNs (Kirkpatrick et al., 2017).

We found that the synaptic noise may play multiple valuable roles in SNNs. On the one hand, it can help in learning new data (see results on noise regularization). On the other hand, the noise can help in determining the appropriate variability range of the weights (as in the Langevin dynamics method). Certain weight variability during learning is vital, as freezing part of the weights may prevent further network training. Also, we found that freezing the largest weights does not prevent SNN from forgetting initial knowledge: the importance of weight does not have a direct relation to its magnitude.

All the CF prevention techniques (architecture-based methods, memory replay methods, and regularization methods) have their benefits. The introduction of lateral inhibition allowed to decorrelate neuronal activities in feature extracting layers. The approach with maintained tiny memory together with regularization technique based on Langevin dynamics demonstrated the superiority among other methods in overcoming catastrophic forgetting. However, current methods could not overcome the forgetting completely, and further research is still needed.

# Acknowledgments

We'd like to acknowledge Dr. M. Mozafari and Dr. T. Masquelier for helpful advice and suggestions. The reported study was funded by the Russian Foundation for Basic Research (project numbers 18-47-732006 and 20-07-00974).

# Supplementary Material: Continuous learning of spiking networks trained with local rules

D.I.Antonov[a], K.V.Sviatov[b], S.Sukhov[a,*]

[a]Kotelnikov Institute of Radio Engineering and Electronics of Russian Academy of Sciences (Ulyanovsk branch), 48/2 Goncharov Str., Ulyanovsk 432011, Russia

[b]Ulyanovsk State Technical University, Severny Venets 32, Ulyanovsk 432027, Russia

Here, we present the behavior of neural network's accuracy in incremental domain learning experiments (see the main text of the paper for more details). Spiking neural network (SNN) was trained on Task 1 (images of digits MNIST) till reaching the classification accuracy of ≈ 93%. After that, the network was trained on Task 2 (images of Latin letters EMNIST) over 100 epochs. During training on Task 2, we tested several methods to prevent the catastrophic forgetting of Task 1 (see the main text for details). Figure S1 shows the result of this training.

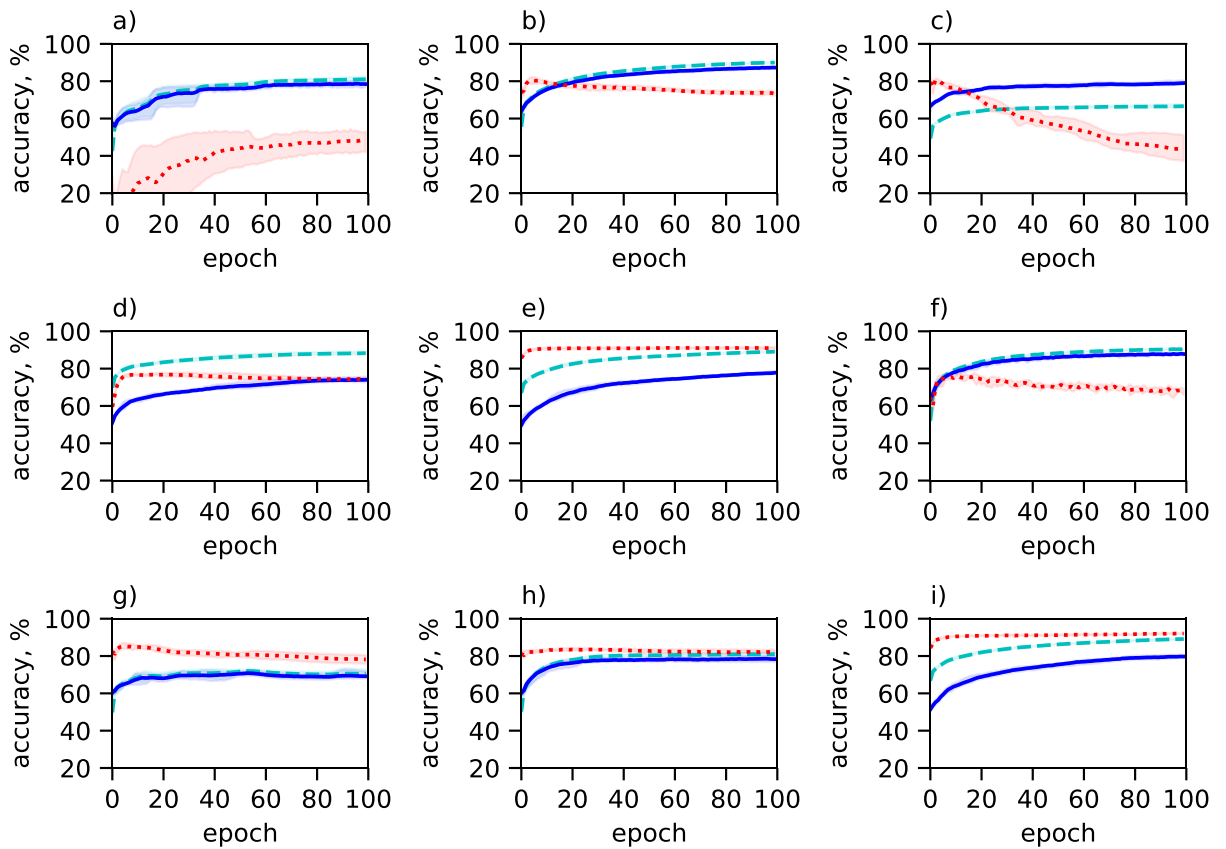

**Figure S1**: The accuracy of SNN in incremental domain learning experiments. Different panels show the efficiency of different methods for catastrophic forgetting prevention: (a) Catastrophic forgetting, (b) Lateral inhibition, (c) Pseudo-rehearsal, (d) Few-short self-remainder (0.25% stored samples), (e) Few-short self-remainder (10% stored samples), (f) Noise regularization, (g) Frozen large weights, (h) Langevin dynamics, (i) Joint training. Lines represent the mean; the shaded areas show the standard deviation. Dotted red lines show the recognition accuracy on Task 1; solid blue lines show the recognition test accuracy on Task 2; dashed cyan lines show the recognition accuracy of the training dataset of Task 2.

* Corresponding author
E-mail address: ssukhov@ulireran.ru (Sergey Sukhov)

The common feature of all the dependencies is the sharp drop in the accuracy for Task 1 at the first step of learning Task 2. This drop is determined by the resetting of learning rates before starting learning Task 2. Interestingly, the accuracy for Task 1 may partially recover during the learning on Task 2 just because of adaptation of learning rates (Figure S1(a)).

The training dataset consists either entirely of dataset of Task 2 (images of letters) (panels a, b, f, g, h) or the mix of this dataset with samples taken from memory buffer (synthetic samples (c), depleted dataset 1 (d, e), or complete dataset of Task 1 (i)).